\PassOptionsToPackage{table}{xcolor}
\documentclass[sigconf,nonacm]{acmart}

\AtBeginDocument{%
  }

\copyrightyear{2026}
\acmYear{2026}
\setcopyright{cc}
\setcctype{by}
\acmDOI{}
\acmISBN{}
\usepackage{multirow}

\newcommand{\best}[1]{\cellcolor[HTML]{DCEFE3}#1} 
\newcommand{\maxc}[1]{\cellcolor[HTML]{FBE4D5}#1} 

\begin{document}

\title{Invisible in Space, Visible in Time: Motion Vision CAPTCHA against GUI Agents}

\author{Zeyu Zhang}
\affiliation{%
  \institution{Shanghai Jiao Tong University}
  \city{Shanghai}
  \country{China}}
\affiliation{%
  \institution{Shanghai Artificial Intelligence Laboratory}
  \city{Shanghai}
  \country{China}}
\email{zeyuzhang@sjtu.edu.cn}

\author{Dingyi Rong}
\affiliation{%
  \institution{Shanghai Jiao Tong University}
  \city{Shanghai}
  \country{China}}
\affiliation{%
  \institution{Shanghai Artificial Intelligence Laboratory}
  \city{Shanghai}
  \country{China}}
\email{r892546826@sjtu.edu.cn}

\author{Zijian Chen}
\affiliation{%
  \institution{Shanghai Jiao Tong University}
  \city{Shanghai}
  \country{China}}
\affiliation{%
  \institution{Shanghai Artificial Intelligence Laboratory}
  \city{Shanghai}
  \country{China}}
\email{zijian.chen@sjtu.edu.cn}

\author{Zicheng Zhang}
\affiliation{%
  \institution{Shanghai Artificial Intelligence Laboratory}
  \city{Shanghai}
  \country{China}}
\email{zzc1998@sjtu.edu.cn}

\author{Xiongkuo Min}
\correspondingauthor
\affiliation{%
  \institution{Shanghai Jiao Tong University}
  \city{Shanghai}
  \country{China}}
\email{minxiongkuo@sjtu.edu.cn}

\author{Guangtao Zhai}
\correspondingauthor
\affiliation{%
  \institution{Shanghai Jiao Tong University}
  \city{Shanghai}
  \country{China}}
\affiliation{%
  \institution{Shanghai Artificial Intelligence Laboratory}
  \city{Shanghai}
  \country{China}}
\email{zhaiguangtao@sjtu.edu.cn}

\renewcommand{\shortauthors}{Zeyu Zhang et al.}
\begin{abstract}
Most existing visual CAPTCHAs remain spatially solvable: the required information is exposed by static appearance, local structure, and interface state. This assumption is weakened by advances in multimodal large language models (MLLMs) and Graphical User Interface (GUI) agents, which exhibit strong visual perception, reasoning, and browser interaction capabilities. We propose \textbf{Motion Vision CAPTCHA (MVCAP)}, a hierarchical motion-based CAPTCHA framework in which target semantics are instantiated as motion-defined foreground structures and become recoverable only through temporal segregation from a dynamically evolving background. Built on this shared principle, MVCAP is instantiated in three perceptually progressive levels: coherent motion, structural motion, and biological motion. To evaluate this framework, we introduce \textbf{MVCAP-Bench}, a browser-based benchmark with 600 live CAPTCHA instances, together with a matched foreground-only control benchmark, \textbf{MVCAP-Bench-FG}. We evaluate humans, Browser Use agents, native computer use agents, and a supplementary offline VQA setting derived from the same instances. Results reveal a substantial human--agent gap: on the full MVCAP-Bench, human accuracy reaches 99.6\%, whereas the best GUI agent achieves only 16.8\%, close to the six-way chance level. The foreground-only control further shows that the key difficulty comes from dynamic background camouflage rather than answer format or browser interaction alone. These findings identify a measurable human--agent perception gap and position MVCAP-Bench as a benchmark for studying motion-defined perception in current agents.
\end{abstract}



\maketitle

\begin{figure*}[t]
  \centering
  \includegraphics[width=\textwidth]{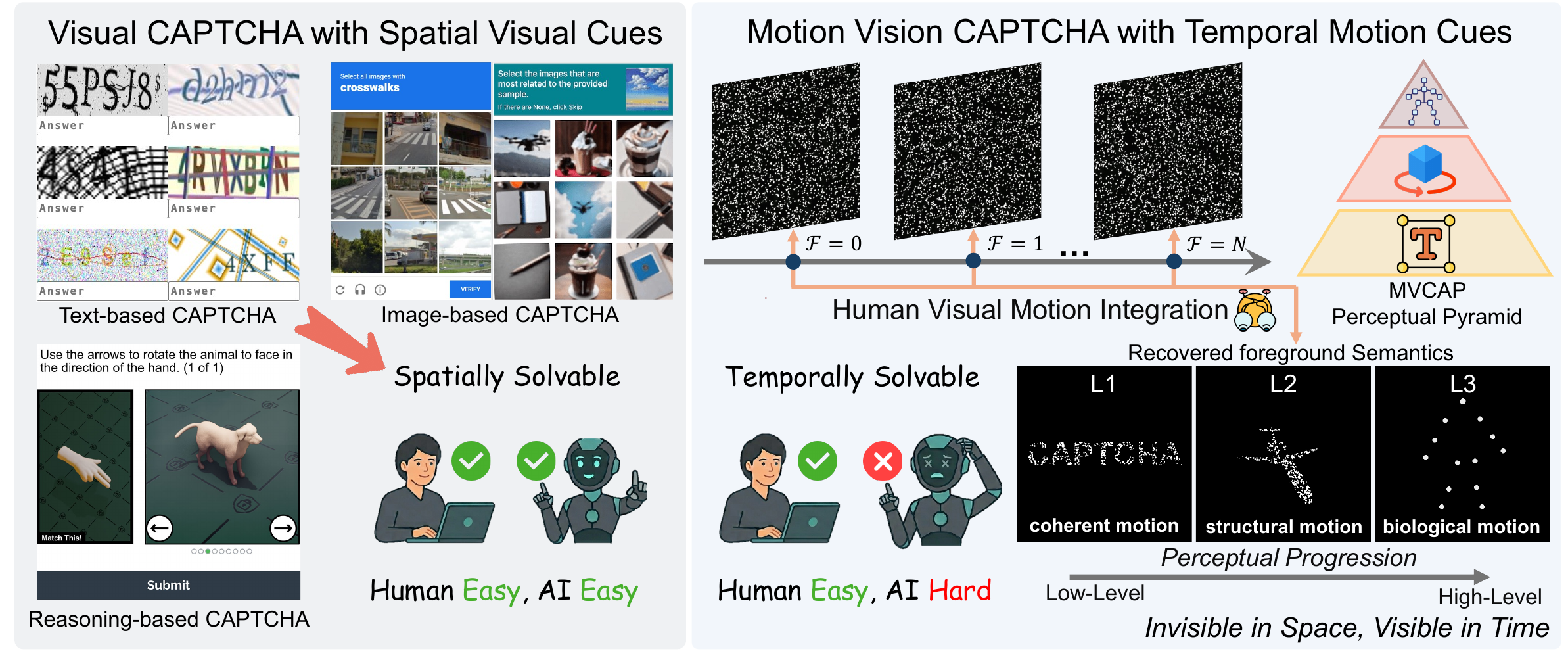}
  \caption{Traditional visual CAPTCHAs are spatially solvable because their semantics are exposed by strong text, image, or reasoning cues. We propose Motion Vision CAPTCHA (MVCAP), a hierarchical motion-based CAPTCHA framework in which target semantics are instantiated as motion-defined foreground structures and become recoverable only through temporal segregation from the background over time. As a result, MVCAP remains easy for humans, but difficult for current AI agents.}
  \Description{A teaser figure comparing conventional visual CAPTCHAs and Motion Vision CAPTCHA. The left side shows text-based, image-based, and reasoning-based CAPTCHAs that reveal strong spatial cues. The right side shows MVCAP, where single frames contain no directly recoverable semantics and the hidden foreground becomes visible only through temporal motion integration. Three levels are illustrated: coherent motion, structural motion, and biological motion. The figure summarizes the observed human--agent gap under the evaluated protocol.}
  \label{teaser}
\end{figure*}

\section{Introduction}

As the cornerstone of web security, CAPTCHAs (Completely Automated Public Turing tests to tell Computers and Humans Apart) serve as essential gatekeepers for protecting multimedia services from malicious automated exploitation\cite{von2003captcha}. Among diverse implementations, visual CAPTCHAs have historically been the predominant deployed paradigm, whereas audio CAPTCHAs are typically used as accessibility fallbacks\cite{bigham2009evaluating}. These challenges ingeniously capitalize on the sophisticated cognitive abilities of the Human Visual System (HVS) by assigning perceptual tasks that remain trivial for humans yet pose significant computational hurdles for traditional automated agents\cite{ferzli2006captcha}.

Over the past decade, visual CAPTCHAs have evolved from text-based and image-based designs to reasoning-based schemes, with challenge formats shifting from simple recognition tasks to more complex multi-step reasoning\cite{zhang2020robust, datta2005imagination, gao2021research}. Representative systems, such as Google reCAPTCHA\cite{von2008recaptcha}, GeeTest CAPTCHA\cite{geetest_v4}, and Arkose Labs’ Arkose MatchKey\cite{arkose_matchkey}, illustrate this trajectory. Despite their apparent diversity, however, these CAPTCHAs largely share a common design assumption: they are primarily spatially solvable, meaning that the information needed to solve a challenge is encoded in static appearance, local visual structure, and the current interface state, rather than in temporal evolution or long-range dependencies. This assumption is increasingly weakened by advances in machine intelligence. Multimodal Large Language models (MLLMs) now exhibit strong visual perception and can reliably extract fine-grained cues, including texture, object appearance, local spatial structure, and semantic relationships\cite{yin2024survey}. Graphical User Interface (GUI) agents further extend this capability with multi-step planning and execution directly on graphical interfaces\cite{hong2024cogagent}. Consequently, spatially solvable CAPTCHA cues are increasingly accessible to current automated agents.

As illustrated in Fig.~\ref{teaser}, we propose \textbf{Motion Vision CAPTCHA (MVCAP)}, a hierarchical motion-based CAPTCHA framework in which target semantics are instantiated as motion-defined foreground structures and are recoverable only through temporal segregation from the background. Unlike conventional CAPTCHAs, which rely mainly on static spatial cues, MVCAP is designed so that target semantics remain inaccessible in isolated frames and become perceptible only through temporal integration of motion cues. By leveraging the human visual system's strong capability for visual motion perception and temporal integration\cite{albright1995visual}, MVCAP tests whether current agents can recover motion-defined semantics that remain accessible to human observers. Under this shared principle, MVCAP is instantiated in three perceptually progressive levels spanning coherent motion, structural motion, and biological motion, covering complementary stages of temporal perception.

Building on the MVCAP framework, we introduce \textbf{MVCAP-Bench}, a browser-based benchmark for evaluating whether agents can solve motion-defined CAPTCHA challenges. MVCAP-Bench contains 600 instances in total, with 200 instances for each of the three levels. We further construct \textbf{MVCAP-Bench-FG}, a matched foreground-only control benchmark that preserves the same semantic targets, prompts, answer options, and browser interface while removing background camouflage. We evaluate humans, Browser Use-based GUI agents such as GPT-5.4\cite{openai_gpt54_2026}, Gemini-3.1-Pro\cite{google_gemini31pro_2026}, Claude-Opus-4.6\cite{anthropic_claude_opus46_2026}, and Qwen3.5-Plus\cite{alibaba_qwen35plus_2026}, native computer-use agents\cite{openai_computer_use_2026, google_computer_use_2026, anthropic_computer_use_2026}, and a supplementary offline VQA setting derived from the same instances. Results reveal a substantial human--agent gap: on the full MVCAP-Bench, human accuracy reaches \textbf{99.6\%}, whereas the best GUI agent achieves only 16.8\%, close to the six-way chance level. Under the evaluated browser-level threat model, these results reveal a measurable gap in recovering motion-defined semantics under dynamic background camouflage.

In summary, our contributions are three-fold:
\begin{itemize}
    \item We propose \textbf{Motion Vision CAPTCHA}, a hierarchical motion-based CAPTCHA framework for human--machine differentiation, in which target semantics are instantiated as motion-defined foreground structures and become recoverable only through temporal segregation from the background. MVCAP is instantiated in three levels: coherent motion, structural motion, and biological motion.
    \item We introduce \textbf{MVCAP-Bench}, a browser-based benchmark with 600 motion-defined CAPTCHA instances organized into three levels, with 200 instances per level, together with a matched foreground-only control benchmark, \textbf{MVCAP-Bench-FG}, for isolating the effect of dynamic background camouflage on motion-defined perception.
    \item We conduct a systematic evaluation of humans and frontier GUI agents across Browser Use, native computer use, and offline VQA settings, and reveal a large performance gap between humans and agents. Humans achieve 99.6\% accuracy on MVCAP-Bench, while the best GUI agent achieves only 16.8\%, close to random guess. Controlled comparisons further show that the main source of difficulty lies in dynamic background camouflage rather than answer format or browser interaction alone.
\end{itemize}

\begin{figure*}[t]
    \centering
    \includegraphics[width=\textwidth]{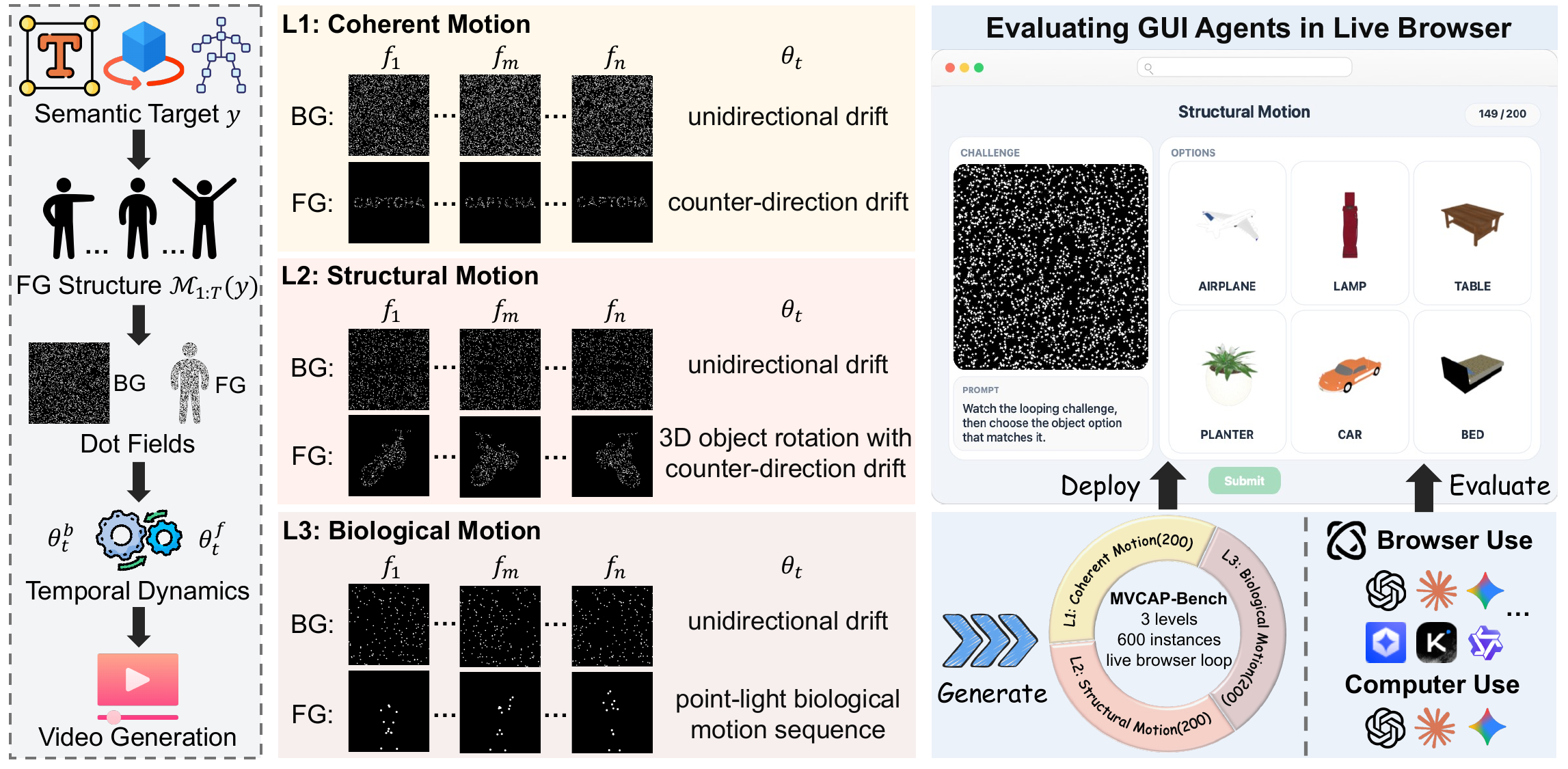}
    \caption{Overview of MVCAP and its browser-based evaluation setting. 
    Left: a semantic target $y$ is instantiated as a temporally evolving foreground structure, embedded into foreground/background dot fields, and rendered as a challenge video. 
    Middle: the shared motion vision principle is instantiated in three levels, coherent motion (L1), structural motion (L2), and biological motion (L3). 
    Right: in MVCAP-Bench, MLLM-based agents solve motion CAPTCHA instances through a live browser interface.}
    \Description{Overview of the Motion Vision CAPTCHA framework and its browser-based benchmark. The left panel shows the generation pipeline from semantic target to foreground structure, dot fields, temporal dynamics, and final video challenge. The middle panel illustrates the three levels of MVCAP: coherent motion, structural motion, and biological motion. The right panel shows the live browser interface used in MVCAP-Bench, where an agent watches a looping motion challenge and selects the matching answer option.}
    \label{fig:l2_browser_example}
\end{figure*}

\section{Related Work}
\subsection{Visual CAPTCHAs}
Visual CAPTCHAs have evolved from distorted text\cite{bursztein2011text} to image recognition\cite{google_recaptcha_v2,luo2025open} and multi-step visual reasoning\cite{ding2025illusioncaptcha}, reflecting a continuing arms race with automated solvers\cite{guerar2021gotta}. Despite their different interfaces and recognition demands, most remain spatially solvable: the answer-relevant evidence is encoded in static appearance, local structure, or the current interface state\cite{deng2025oedipus}. Animated and video CAPTCHAs introduce temporal presentation or interaction\cite{athanasopoulos2006enhanced,kluever2009balancing}, but animation alone does not make temporal integration necessary; some schemes can still be attacked through information recovered from extracted frames\cite{nguyen2012breaking}. MVCAP therefore differs not in using video as the delivery medium, but in making the target itself a motion-defined foreground embedded in dynamic background camouflage. Solving it requires temporal foreground--background segregation rather than recognizing evidence already exposed in an individual frame.

\subsection{MLLMs and GUI Agents}

MLLMs have improved visual grounding and reasoning\cite{zhang2025puzzlebench,ma2024groma}. When used as GUI-agent backbones, these capabilities support screenshot understanding, web reasoning, and multi-step interaction; VisualWebArena, WebVoyager, and OSWorld document this progress across browser and computer tasks\cite{koh2024visualwebarena,he2024webvoyager,xie2024osworld}. CAPTCHA-focused evaluations further show that agentic vision--language models generalize across diverse visual challenges\cite{teoh2025captchas}, while Open CaptchaWorld and COGNITION demonstrate non-trivial success on real-world CAPTCHA tasks, especially those dominated by spatial perception and shallow interaction\cite{luo2025open,wang2025cognition}. These advances weaken the assumptions behind CAPTCHAs whose decisive cues are statically exposed. MVCAP instead tests whether current GUI agents can recover motion-defined semantics when the target must first be segregated from dynamic camouflage.

\subsection{Motion-Defined Perception}

Human vision can recover meaningful structure from motion rather than from spatial appearance alone. Coherent-motion grouping supports foreground segregation, form-from-motion reveals object structure, and sparse biological motion conveys action semantics\cite{williams1984coherent,grossman1999perception,vaina2001functional}. Current multimodal models do not yet show the same reliability in these settings. Evaluated video-language models perform poorly when semantic information is carried primarily by temporal patterns\cite{upadhyay2026time}, and related limitations appear in biological-motion recognition\cite{chen2025can}. MVCAP operationalizes this perceptual contrast through coherent, structural, and biological motion challenges. Its purpose is to measure the resulting motion-perception gap in current agents, not to claim permanent machine incapability.

\section{Threat Model}

\noindent \textbf{Attacker Goals.} We consider a realistic browser-based attacker whose objective is to autonomously solve MVCAP challenges and thereby bypass browser-based human verification without human intervention. Successful attacks may enable unauthorized automated access to protected web services, weakening the intended distinction between legitimate human users and automated agents.

\noindent \textbf{Attacker Capabilities.} We model the attacker as an MLLM-based GUI agent operating in a live browser environment. Following prior GUI agent settings, the interaction is formulated as a partially observable Markov decision process (POMDP),
\begin{equation}
\mathcal{E} = \langle \mathcal{S}, \mathcal{O}, \mathcal{A}, \mathcal{T} \rangle,
\end{equation}
where $\mathcal{S}$ denotes the hidden session-state space, $\mathcal{O}$ denotes the observation space, $\mathcal{A}$ denotes the space of the GUI action space, and $\mathcal{T}$ denotes the environment transition function. At step $t$, the agent receives an observation $o_t \in \mathcal{O}$, which may include rendered screenshots, visible prompt text, candidate options, and structured interface representations such as DOM or accessibility information. Based on this observation, the agent selects an action $a_t \in \mathcal{A}$, such as clicking interface elements, choosing one of the six options, or submitting an answer. The environment then evolves according to
\begin{equation}
s_{t+1} \sim \mathcal{T}(s_t, a_t),
\end{equation}
and the agent follows a policy
\begin{equation}
a_t \sim \pi_\theta(o_{1:t}, a_{1:t-1}),
\end{equation}
where $\pi_\theta$ denotes an MLLM-based decision policy parameterized by $\theta$. Under this threat model, the attacker is allowed to access the live webpage, observe rendered browser content, and employ Browser Use tools together with multi-step reasoning, but must solve the challenge through iterative GUI interaction rather than offline static classification.

\section{Motion Vision CAPTCHA}

\subsection{Overview}

As summarized in Fig.~\ref{fig:l2_browser_example}, MVCAP is a hierarchical motion-based CAPTCHA framework grounded in the human visual system's ability to recover meaningful structure from visual motion over time. Unlike traditional visual CAPTCHAs, which are mainly solvable from static spatial cues, MVCAP is designed such that target semantics remain effectively inaccessible in isolated frames and become perceptible only through temporal integration of motion cues. By leveraging this remarkable and distinctive strength of human visual motion perception, MVCAP shifts CAPTCHA solving from static recognition to temporally grounded semantic perception.

Built on this shared principle, MVCAP is instantiated in three explicit hierarchical motion levels: L1 coherent motion, L2 structural motion, and L3 biological motion. These levels form a perceptual progression from motion-grouped text perception, to motion-defined 3D object structure perception, and finally to action perception from point-light biological motion. Across all levels, target semantics are instantiated as temporally evolving foreground structures and embedded into backgrounds that suppress reliable recovery from static spatial cues alone.

\subsection{Shared Motion Vision Principle}

Let $y$ denote the target semantics of a CAPTCHA instance, such as a text string, an object category, or an action category. MVCAP first instantiates $y$ as a temporally evolving foreground structure:
\begin{equation}
M_{1:T}(y) = \{M_t(y)\}_{t=1}^{T},
\end{equation}
where $M_t(y)$ denotes the foreground structure at time step $t$. The exact form of $M_t(y)$ is level-dependent: it may be a text-shaped support region, a motion-defined 3D object structure, or a point-light articulated human motion pattern.

MVCAP then samples a foreground dot field $A_f$ and a background distractor dot field $A_b$. The observed challenge sequence $X_{1:T} = \{X_t\}_{t=1}^{T}$ is generated frame by frame as
\begin{equation}
X_t = \mathcal{C}\big(\mathcal{R}_f(M_t(y), A_f, \theta_t^f), \mathcal{R}_b(A_b, \theta_t^b)\big),
\end{equation}
where $\mathcal{R}_f(\cdot)$ and $\mathcal{R}_b(\cdot)$ denote level-specific rendering operators, and $\theta_t^f$ and $\theta_t^b$ parameterize the temporal dynamics of the foreground and background, respectively. The resulting sequence is rendered and presented as a challenge video.

In this formulation, $M_t(y)$ specifies the semantic foreground structure, while $\theta_t^f$ and $\theta_t^b$ determine how the foreground and background evolve over time. Importantly, the background is not a static noise source, but a temporally evolving distractor field with its own motion dynamics. Concretely, in L1 the foreground and background are separated primarily by opposite drift directions; in L2 the semantic structure evolves through 3D object rotation while both dot fields additionally undergo differential drift; and in L3 the foreground evolves as a point-light biological motion sequence against a drifting sparse distractor field. MVCAP therefore does not merely corrupt the target with added noise, but conceals it through competing foreground and background motion over time.

Although the rendering details differ across levels, the underlying principle remains the same: target semantics are not designed to be reliably recoverable from isolated frames, but instead become perceptible through temporal integration of differential motion between foreground and background. In dense settings such as L1 and L2, the rendered foreground and background appear as visually similar dense dot fields with different temporal dynamics. In sparse settings such as L3, the foreground is a point-light biological motion pattern and the background is a sparse distractor point field. Under all three instantiations, MVCAP preserves the same motion vision principle: semantics are weak or inaccessible in static space, but emerge over time through motion-defined perception.

\begin{table*}[!t]
\centering
\scriptsize
\setlength{\tabcolsep}{2.0pt}
\renewcommand{\arraystretch}{1.10}
\caption{Comparison with representative CAPTCHA benchmarks and systems. MVCAP-Bench differs from prior work by jointly combining browser-based evaluation, procedurally generated instances, and motion-defined challenges that are not solvable from static spatial cues alone.}
\label{tab:captcha_benchmark_comparison}

\resizebox{\textwidth}{!}{%
\begin{tabular}{
>{\raggedright\arraybackslash}m{2.6cm}
>{\centering\arraybackslash}m{1.7cm}
>{\centering\arraybackslash}m{2.0cm}
>{\centering\arraybackslash}m{1.6cm}
>{\centering\arraybackslash}m{3.1cm}
>{\centering\arraybackslash}m{1.6cm}
>{\centering\arraybackslash}m{1.6cm}
>{\centering\arraybackslash}m{1.4cm}}
\toprule
\textbf{Work} &
\textbf{Year/Venue} &
\textbf{Modality} &
\shortstack{\textbf{Evaluation} \\ \textbf{Setting}} &
\shortstack{\textbf{Evaluation} \\ \textbf{Size}} &
\shortstack{\textbf{Procedural} \\ \textbf{Generation}} &
\shortstack{\textbf{Spatially} \\ \textbf{Solvable}} &
\shortstack{\textbf{Human} \\ \textbf{Pass (\%)}} \\
\midrule

Diff-CAPTCHA~\cite{jiang2023diff}
& 2023 / arXiv
& image + text
& Offline
& 1000 instances
& Yes
& Yes
& 93.5 \\

Aura-CAPTCHA~\cite{chandra2025aura}
& 2025 / arXiv
& image + text + audio
& Online
& N/R
& Yes
& Partial
& 92.8 \\

IllusionCAPTCHA~\cite{ding2025illusioncaptcha}
& 2025 / WWW
& image + text
& Offline
& 30 instances
& Yes
& Yes
& 87.0 \\

Open CaptchaWorld~\cite{luo2025open}
& 2025 / NeurIPS
& image + text
& Browser
& 225 instances / 20 types
& No
& Yes
& 93.3 \\

Spatial CAPTCHA~\cite{kharlamova2025spatial}
& 2026 / ICLR
& image + text
& Offline
& 1050 instances / 7 tasks
& Yes
& Yes
& 99.8 \\

\midrule
\rowcolor{black!8}
\textbf{MVCAP-Bench (ours)}
& \textbf{2026 / --}
& \textbf{video + image + text}
& \textbf{Browser}
& \textbf{600 instances / 3 levels / 200 each}
& \textbf{Yes}
& \textbf{No}
& \textbf{99.6} \\
\bottomrule
\end{tabular}
}
\end{table*}

\subsection{Level Instantiations}

\subsubsection{L1: Coherent Motion}

L1 instantiates the target semantics as text. Given a target string $y$, MVCAP constructs a text-shaped foreground structure whose identity remains fixed over time, while the foreground and background follow distinct motion patterns. In our benchmark, the L1 targets are procedurally generated four-character strings rendered as binary text masks. The resulting challenge sequence therefore contains a motion-defined text target embedded in a visually similar dot background. Although the text is weak or ambiguous in isolated frames, coherent motion causes the character structure to become perceptually grouped over time. L1 therefore targets text recognition through coherent motion.

\subsubsection{L2: Structural Motion}

L2 instantiates the target semantics as a motion-defined 3D object structure. Given a target object category $y$, MVCAP constructs a temporally evolving foreground by rendering a 3D object under controlled rotation. In our benchmark, the L2 object categories and meshes are derived from \textit{ShapeNetCore}~\cite{chang2015shapenet} using a curated set of object classes and selected 3D assets. Embedded into a visually similar dot background, the evolving object structure is difficult to recover from any single frame alone, but becomes perceptible over time through temporal integration. L2 therefore targets object recognition through structural motion.

\subsubsection{L3: Biological Motion}

L3 instantiates the target semantics as biological motion. Given a target action category $y$, MVCAP constructs a sparse articulated foreground sequence using a point-light display of human motion. In our benchmark, the L3 action sequences are derived from \textit{BABEL} annotations aligned with local \textit{CMU} motion-capture data~\cite{punnakkal2021babel,cmu_mocap}, and are converted into point-light biological motion stimuli for the challenge videos while simultaneously rendering the corresponding answer options. Unlike L1 and L2, which use denser dot carriers, L3 combines a sparse point-light foreground with sparse background distractors. The action pattern thus becomes perceptible through temporally structured motion cues rather than static appearance alone. L3 therefore targets action recognition through biological motion.

Additional details are provided in the supplementary material.
\subsection{Perceptual Progression Across Levels}

The three levels of MVCAP are designed to engage different levels of motion perception in the human visual system, rather than to represent a simple increase in task difficulty. L1 targets coherent motion perception, where humans group motion-defined text from background noise. L2 targets structural motion perception, where humans recover object structure from temporally evolving 3D motion. L3 targets biological motion perception, where humans recognize actions from sparse articulated motion cues. Taken together, the three levels form a perceptual progression from coherent motion $\rightarrow$ structural motion $\rightarrow$ biological motion.

\subsection{Benchmark Construction}

To evaluate whether MLLM-based agents can solve motion-defined CAPTCHAs in realistic interactive settings, we construct MVCAP-Bench, a browser-based benchmark built on the three-level MVCAP framework. Each instance is formulated as a page-level multimodal VQA task consisting of a looping challenge video, a textual prompt, and six answer options, from which the agent must select and submit the correct answer in a live browser interface. Each challenge video is 3 seconds long and rendered at 30 frames per second. MVCAP-Bench contains 600 instances organized into three levels, with 200 instances per level, corresponding to coherent motion, structural motion, and biological motion. These levels are designed to engage different levels of motion perception in the human visual system, rather than to define a simple progression in human difficulty. In addition, we construct MVCAP-Bench-FG, a matched variant in which the challenge videos retain only the foreground content while keeping the semantic targets, prompts, answer options, and browser interface unchanged, thereby providing a controlled reference for isolating the role of dynamic background camouflage and motion-based concealment. As summarized in Table~\ref{tab:captcha_benchmark_comparison}, MVCAP-Bench differs from prior CAPTCHA benchmarks and systems by jointly combining browser-based evaluation, procedurally generated instances, and motion-defined challenges that are not solvable from static spatial cues alone. Detailed generation parameters and evaluation settings are provided in the supplementary material.

\begin{table*}[t]
\centering
\scriptsize
\setlength{\tabcolsep}{2.5pt}
\renewcommand{\arraystretch}{1.1}
\caption{Main results on MVCAP-Bench and MVCAP-Bench-FG. We compare three types of methods: (1) Browser Use instantiated with ten multimodal backbones; (2) native computer use agents based on GPT-5.4, Claude-Sonnet-4.6, and Gemini-3-Flash; (3) two baselines, random guess and human level. Each method is reported by three levels with Pass@1 and cost (\$), together with average Pass@1 and total cost across levels. {\color[HTML]{DCEFE3}\rule{0.9em}{0.9em}} denotes highest Pass@1; {\color[HTML]{FBE4D5}\rule{0.9em}{0.9em}} denotes highest cost.}
\label{tab:browser_use_main_ablation}

\resizebox{\textwidth}{!}{%
\begin{tabular}{>{\raggedright\arraybackslash}m{2cm}cccccccccccccccc}
\toprule
\multirow[c]{3}{2cm}[-1.5ex]{\raggedright\arraybackslash\textbf{Method}}
& \multicolumn{8}{c}{\textbf{MVCAP-Bench}} & \multicolumn{8}{c}{\textbf{MVCAP-Bench-FG}} \\
\cmidrule(lr){2-9} \cmidrule(lr){10-17}
& \multicolumn{2}{c}{\textbf{L1}} & \multicolumn{2}{c}{\textbf{L2}} & \multicolumn{2}{c}{\textbf{L3}} & \multicolumn{2}{c}{\textbf{Summary}}
& \multicolumn{2}{c}{\textbf{L1}} & \multicolumn{2}{c}{\textbf{L2}} & \multicolumn{2}{c}{\textbf{L3}} & \multicolumn{2}{c}{\textbf{Summary}} \\
\cmidrule(lr){2-7} \cmidrule(lr){8-9} \cmidrule(lr){10-15} \cmidrule(lr){16-17}
& Pass@1 & Cost & Pass@1 & Cost & Pass@1 & Cost & Avg. & Total
& Pass@1 & Cost & Pass@1 & Cost & Pass@1 & Cost & Avg. & Total \\
\midrule
\rowcolor{black!8}\multicolumn{17}{l}{\textit{Browser Use}} \\
GPT-5-mini & 13.5 & \$0.72 & 13.5 & \$0.90 & \best{20.5} & \$0.93 & 15.8 & \$2.56 & 86.5 & \$0.90 & 45.0 & \$1.03 & 16.5 & \$0.96 & 49.3 & \$2.89 \\
GPT-5.4 & \best{17.0} & \$9.35 & 16.5 & \maxc{\$26.00} & 17.0 & \maxc{\$21.72} & \best{16.8} & \$57.06 & 96.5 & \$16.08 & 66.5 & \$15.31 & 22.0 & \maxc{\$22.38} & 61.7 & \$53.77 \\
Claude-Sonnet-4-6 & 13.5 & \$16.17 & \best{17.0} & \$11.35 & 16.5 & \$12.08 & 15.7 & \$39.60 & 96.5 & \$12.36 & 55.5 & \$12.04 & 21.5 & \$11.78 & 57.8 & \$36.18 \\
Claude-Opus-4-6 & 14.0 & \maxc{\$20.63} & 14.0 & \$19.78 & 18.0 & \$19.56 & 15.3 & \maxc{\$59.96} & 97.0 & \maxc{\$20.70} & 45.5 & \maxc{\$20.38} & 21.5 & \$21.25 & 54.7 & \maxc{\$62.33} \\
Gemini-3-Flash & 13.5 & \$1.16 & 13.5 & \$1.16 & \best{20.5} & \$2.34 & 15.8 & \$4.66 & \best{100.0} & \$1.33 & 67.0 & \$1.09 & 22.0 & \$1.21 & 63.0 & \$3.63 \\
Gemini-3.1-Pro & 13.5 & \$3.96 & 13.5 & \$4.00 & 14.5 & \$4.25 & 13.8 & \$12.21 & 99.0 & \$4.67 & \best{75.5} & \$4.23 & \best{29.5} & \$4.71 & \best{68.0} & \$13.61 \\
Doubao-Seed-2.0-Pro & 16.5 & \$2.85 & 14.0 & \$4.04 & 18.5 & \$1.44 & 16.3 & \$8.33 & 96.0 & \$1.47 & 50.0 & \$1.46 & 15.0 & \$1.36 & 53.7 & \$4.29 \\
GLM-4.6V & 13.5 & \$1.87 & 14.0 & \$1.69 & 16.0 & \$1.62 & 14.5 & \$5.19 & 79.0 & \$1.82 & 22.5 & \$1.68 & 17.5 & \$1.69 & 39.7 & \$5.19 \\
Kimi-K2.5 & 13.5 & \$3.11 & 16.5 & \$3.21 & 15.0 & \$5.37 & 15.0 & \$11.69 & 86.5 & \$3.06 & 40.0 & \$3.01 & 13.0 & \$6.26 & 46.5 & \$12.33 \\
Qwen3.5-Plus & 13.5 & \$0.45 & 13.5 & \$0.44 & 17.5 & \$0.63 & 14.8 & \$1.52 & 98.0 & \$0.47 & 48.5 & \$0.46 & 15.0 & \$0.72 & 53.8 & \$1.65 \\
\midrule
\rowcolor{black!8}\multicolumn{17}{l}{\textit{Computer Use}} \\
GPT-5.4 & \best{20.0} & \$10.44 & \best{20.0} & \maxc{\$18.43} & \best{22.5} & \maxc{\$20.68} & \best{20.8} & \maxc{\$49.55} & 99.0 & \maxc{\$14.96} & 67.5 & \maxc{\$16.33} & 19.5 & \maxc{\$23.41} & 62.0 & \maxc{\$54.70} \\
Claude-Sonnet-4.6 & 15.0 & \maxc{\$12.86} & 12.5 & \$12.41 & 18.0 & \$11.50 & 15.2 & \$36.77 & \best{100.0} & \$11.50 & 59.0 & \$12.84 & \best{23.0} & \$12.32 & 60.7 & \$36.65 \\
Gemini-3-Flash & 15.5 & \$4.61 & 15.5 & \$7.40 & 16.0 & \$6.43 & 15.7 & \$18.44 & \best{100.0} & \$7.50 & \best{72.5} & \$4.61 & 21.7 & \$4.25 & \best{64.7} & \$16.36 \\
\midrule
\rowcolor{black!8}\multicolumn{17}{l}{\textit{Baseline}} \\
Random Guess & 16.7 & - & 16.7 & - & 16.7 & - & 16.7 & - & 16.7 & - & 16.7 & - & 16.7 & - & 16.7 & - \\
Human Level & \best{100.0} & - & \best{100.0} & - & \best{98.8} & - & \best{99.6} & - & \best{100.0} & - & \best{100.0} & - & \best{100.0} & - & \best{100.0} & - \\
\bottomrule
\end{tabular}
}
\end{table*}

\section{Experiment}
\subsection{Experimental Setup}

\noindent\textbf{Agent frameworks.}
We evaluate MVCAP-Bench with three agent frameworks. Our primary framework is \textbf{Browser Use}\cite{browser_use2024}, instantiated with ten multimodal backbones: GPT-5-mini and GPT-5.4\cite{openai_gpt5mini_2026,openai_gpt54_2026}; Claude-Sonnet-4.6 and Claude-Opus-4.6\cite{anthropic_claude_sonnet46_2026,anthropic_claude_opus46_2026}; Gemini-3-Flash and Gemini-3.1-Pro\cite{google_gemini3flash_2025,google_gemini31pro_2026}; GLM-4.6V\cite{zai_glm46v_2025}; Kimi-K2.5\cite{moonshot_kimik25_2026}; Doubao-Seed-2.0-pro\cite{bytedance_seed20pro_2026}; and Qwen3.5-Plus\cite{alibaba_qwen35plus_2026}.

We further evaluate models with native computer use capability, including GPT-5.4, Claude-Sonnet-4.6, and Gemini-3-Flash. Across all settings, agents operate directly in the browser and must solve each instance through screenshot-based observation and GUI interaction alone, without access to any ground-truth target information or auxiliary structured cues beyond the rendered webpage itself.

\noindent\textbf{Benchmarks.}
We conduct experiments on two matched benchmarks: MVCAP-Bench and MVCAP-Bench-FG. MVCAP-Bench is the full benchmark with both foreground and background motion components, while MVCAP-Bench-FG retains only the foreground challenge content and keeps the semantic targets, prompts, options, and browser interface unchanged. The full benchmark contains 600 instances organized into three levels, with 200 instances per level.

\noindent\textbf{Task and baselines.}
Each instance is formulated as a live browser page-level multimodal VQA task containing a looping challenge video, a textual prompt, and six answer options, from which the agent must select and submit one answer. We include two baselines: human and random guess. For the human baseline, we recruit 30 participants and evaluate them on balanced subsets of MVCAP-Bench and MVCAP-Bench-FG, each constructed by sampling 20 instances per level, resulting in 60 instances per dataset.

\noindent\textbf{Metrics.}
We evaluate all methods using \textbf{Pass@1}, where each instance is attempted once and counted as successful only if the submitted answer is correct. We also record the \textbf{monetary cost} of each model during evaluation. Pass@1 measures end-to-end task success under a single browser interaction attempt, while cost reflects the practical expense of deploying MLLM-based agents in live browser settings. Detailed prompting configurations, browser settings, and implementation details are provided in the supplementary material.

\subsection{Experiment Results}

\noindent\textbf{Measured Human--Agent Gap.}
The full benchmark produces a measured gap between humans and the evaluated Browser Use agents. Agent averages range from 13.8\% to 16.8\%, close to the 16.7\% six-way chance level; even the best result, obtained by GPT-5.4, is 16.8\%, whereas humans reach 99.6\%. The full benchmark also compresses the spread among agent backbones: model differences visible on MVCAP-Bench-FG largely disappear after dynamic background camouflage is restored. This gap is consistent across levels rather than being driven by one unusually difficult subset. Human accuracy is 100.0\% on L1 and L2 and 98.8\% on L3, while no Browser Use agent exceeds 20.5\% on any full-benchmark level. Within these configurations, the convergence indicates a shared difficulty in recovering temporally distributed motion cues, rather than choosing among six options presented in the same interface.

\begin{figure*}[t]
    \centering
    \includegraphics[width=0.9\textwidth]{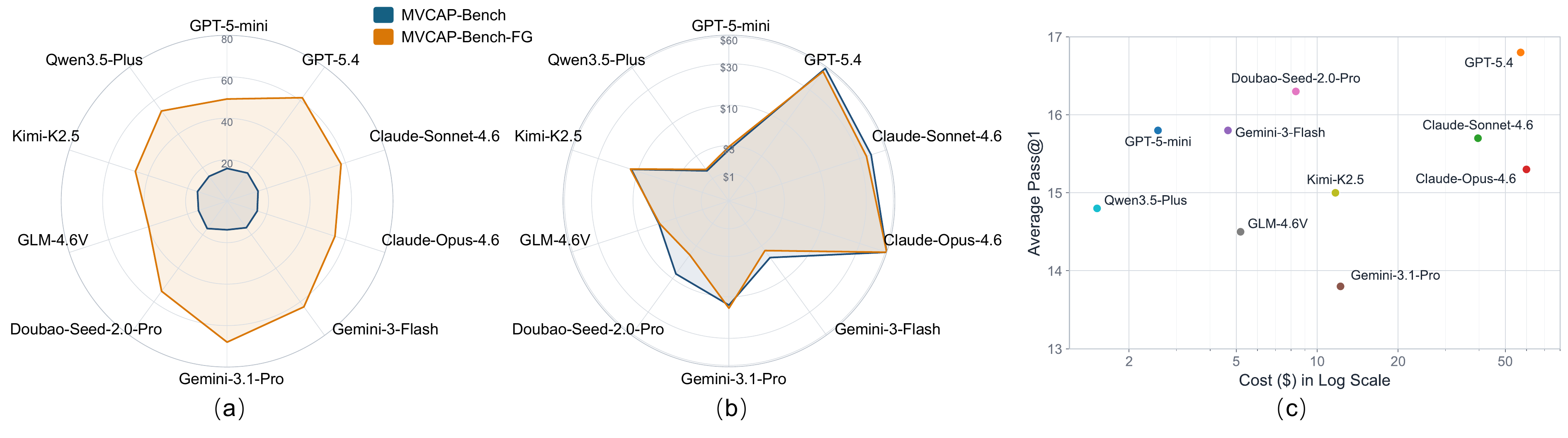}
    \caption{Comparison of Browser Use agents on MVCAP-Bench and MVCAP-Bench-FG. 
    (a) Average Pass@1 across L1--L3. 
    (b) Total cost across L1--L3. 
    (c) Accuracy-cost trade-off on MVCAP-Bench. 
    In (a) and (b), {\color[HTML]{1F4E79}\rule{0.9em}{0.9em}} denotes MVCAP-Bench and {\color[HTML]{D97706}\rule{0.9em}{0.9em}} denotes MVCAP-Bench-FG. 
    The cost radar in (b) uses a log-scaled radial axis for readability, and the x-axis of (c) is shown in log scale.}
    \Description{A combined three-panel figure comparing Browser Use agents. Panel (a) shows average Pass@1 across L1 to L3 for MVCAP-Bench and MVCAP-Bench-FG. Panel (b) shows total cost across L1 to L3 for the same methods, with a log-scaled radial axis. Panel (c) shows the accuracy-cost trade-off on MVCAP-Bench, where each point represents one model and the x-axis is displayed in log scale.}
    \label{fig:browser_use_analysis}
\end{figure*}

\noindent\textbf{Dynamic Background Camouflage Drives the Observed Gap.}
MVCAP-Bench-FG isolates this effect by preserving the semantic targets, prompts, answer options, and browser interface while removing dynamic background camouflage. Average \textsc{Pass@1} increases from 16.8\% to 61.7\% for GPT-5.4, from 13.8\% to 68.0\% for Gemini-3.1-Pro, and from 15.8\% to 63.0\% for Gemini-3-Flash; these correspond to absolute gains of 44.9, 54.2, and 47.2 percentage points, respectively. In the control, the foreground is directly exposed, whereas the full benchmark requires the same foreground to be segregated from a competing dynamic dot field. Because the answer vocabulary and interaction sequence are unchanged, the gains measure the effect of making the motion-defined foreground directly recoverable rather than altering the downstream decision. The paired comparison therefore separates foreground recognition from foreground recovery under camouflage and identifies the latter as a major source of difficulty for the evaluated agents. Similar costs across conditions make reduced effort or early stopping unlikely explanations for the observed performance gap.

\noindent\textbf{The Three Levels Probe Different Perceptual Bottlenecks.}
Table~\ref{tab:browser_use_main_ablation} reveals distinct bottlenecks across motion segregation, structure recovery, and biological-motion interpretation, rather than a uniform easy-to-hard progression. On L1, seven Browser Use backbones reach 96.0\%--100.0\% on MVCAP-Bench-FG, while every backbone remains between 13.5\% and 17.0\% on the full benchmark. This contrast shows that models can recognize the exposed text patterns but fail to segregate them reliably under dynamic camouflage. On L2, the best result rises from 17.0\% to 75.5\%, indicating a strong camouflage effect alongside a remaining challenge in recovering object-level structure from motion. L3 behaves differently: even after foreground isolation, the best result is only 29.5\%. For the evaluated models, this smaller gain is consistent with an additional action-semantic bottleneck when interpreting sparse articulated motion from point-light displays as an action category.

\noindent\textbf{Cost does not predict success.}
Costs vary widely on MVCAP-Bench, yet average \textsc{Pass@1} stays between 13.8\% and 16.8\%. Qwen3.5-Plus costs \$1.52 and obtains 14.8\%, while GPT-5.4 and Claude-Opus-4.6 cost \$57.06 and \$59.96 but reach only 16.8\% and 15.3\%, respectively. The ranking also changes when the foreground is exposed: Gemini-3.1-Pro leads MVCAP-Bench-FG at 68.0\% with a cost of \$13.61, ahead of GPT-5.4 at 61.7\% and Claude-Opus-4.6 at 54.7\%. The scatter plot in Fig.~\ref{fig:browser_use_analysis}(c) therefore shows no monotonic cost--accuracy relation. Higher inference spending alone does not resolve the observed difficulty in these configurations.

\noindent\textbf{The limitation persists across agent embodiments.}
The pattern appears with both Browser Use and native computer use. On MVCAP-Bench, GPT-5.4 scores 16.8\% with Browser Use and 20.8\% with native computer use. Claude-Sonnet-4.6 scores 15.7\% with Browser Use and 15.2\% natively, while Gemini-3-Flash scores 15.8\% with Browser Use and 15.7\% natively. Native computer use is four percentage points higher for GPT-5.4, whereas the two scores differ by less than one point for Claude-Sonnet-4.6 and Gemini-3-Flash. The foreground-only results are similarly stable: GPT-5.4 scores 61.7\% with Browser Use and 62.0\% natively; Claude-Sonnet-4.6 scores 57.8\% with Browser Use and 60.7\% natively; and Gemini-3-Flash scores 63.0\% with Browser Use and 64.7\% natively. Removing camouflage improves every backbone by more than 40 percentage points under either embodiment. With backbones and instances aligned, camouflage removal produces the larger and more consistent shift, indicating that the limitation is not specific to one agent framework but persists across both evaluated control stacks.

\subsection{Case Study}

\begin{figure}[t]
  \centering
  \includegraphics[width=\columnwidth]{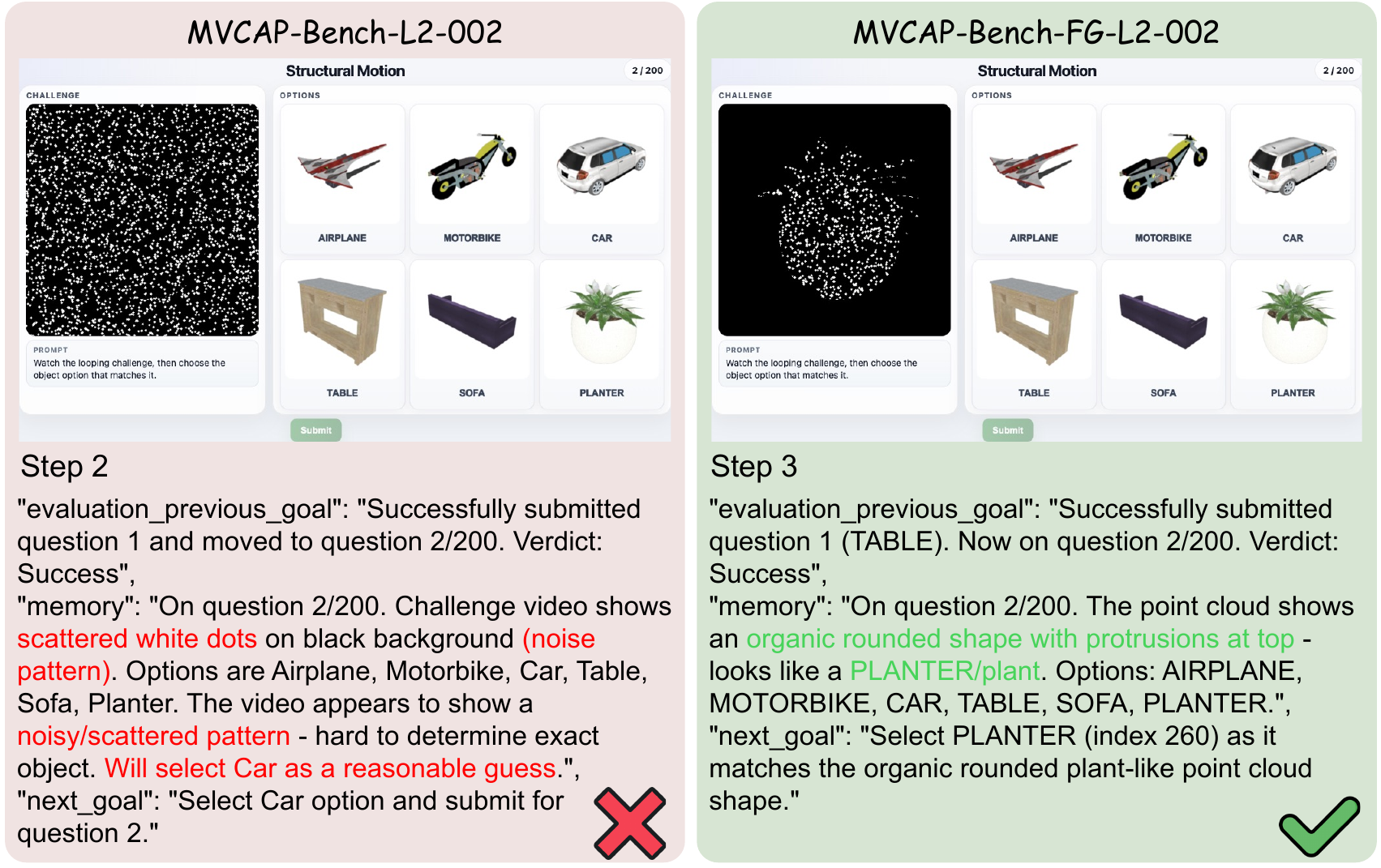}
  \caption{Case study on L2--002 with Claude-Sonnet-4.6. Left: In MVCAP-Bench, the noisy full-scene dot pattern obscures the target object and causes an incorrect guess. Right: In MVCAP-Bench-FG, the clearer foreground allows the model to infer the planter-like shape and select the correct answer.}
  \Description{A two-panel case study for L2-002 with Claude-Sonnet-4.6. In MVCAP-Bench, the full-scene dot pattern obscures the target object and causes an incorrect guess. In MVCAP-Bench-FG, the clearer foreground structure allows the model to recognize the correct answer.}
  \label{fig:case_l2_planter}
\end{figure}

To examine the effect of camouflage on a paired sample, we analyze a representative L2 example in Fig.~\ref{fig:case_l2_planter}. Claude-Sonnet-4.6 fails on MVCAP-Bench but succeeds on MVCAP-Bench-FG for the same underlying target.

In the full benchmark, the model describes the input as noisy and scattered and predicts \textit{car} instead of \textit{planter}. Its trace does not recover a stable object-level structure from the cluttered observation, placing the error before the final answer selection in this example.

In MVCAP-Bench-FG, by contrast, the same semantic target is preserved while the distracting background is removed. Under this setting, the same model is able to identify meaningful structural cues and correctly predicts \textit{planter}. The contrast shows that the success of the foreground-only variant comes from the recovery of perceptually relevant foreground structure, not from a change in task semantics or answer format.

For this paired case, the outcome changes with foreground visibility while the semantic target and answer options remain fixed. The trace is therefore consistent with camouflage obstructing recovery of the relevant object-level evidence for the evaluated model. Additional case studies are provided in the supplementary material.

\subsection{Offline VQA Experiment}

Our primary evaluation is conducted in a live browser environment, where GUI agents must solve MVCAP instances through screenshot-based observation and interaction. This setting is important because it reflects the realistic deployment scenario targeted by MVCAP-Bench. At the same time, however, browser-based evaluation entangles several factors beyond motion understanding itself, including interface perception, action selection, and end-to-end execution. As a result, low performance in the browser setting may arise either from failures in motion-based perception or from additional difficulties introduced by the interaction loop.

To disentangle these factors, we conduct a supplementary offline VQA experiment on both MVCAP-Bench and its foreground-only variant, MVCAP-Bench-FG. Starting from the original browser-based instances, we convert each challenge into a multimodal video question answering sample while preserving the same target semantics and recognition goal. The key difference is that the browser interaction process is removed, so that the model is evaluated more directly on its ability to recover the motion-defined semantics encoded in the Motion Vision CAPTCHA.
For each challenge video, we uniformly sample 10 frames per second and use the sampled frames as the visual input to the VQA model. The model is then asked to answer the corresponding multiple-choice question by selecting the correct option from six candidates. We use accuracy as the evaluation metric. The detailed prompting format and frame sampling configuration are provided in the supplementary material.

\begin{table}[t] 
\centering
\scriptsize
\setlength{\tabcolsep}{5pt}
\renewcommand{\arraystretch}{1.10}
\caption{Results of the \textbf{Offline VQA Experiment} on MVCAP-Bench and MVCAP-Bench-FG. Compared with browser-based evaluation, this setting removes GUI interaction and evaluates models more directly on motion-temporal understanding from sampled video frames. We report accuracy for each level (L1--L3) and the overall average; {\color[HTML]{DCEFE3}\rule{0.9em}{0.9em}} denotes the highest score in each column.}
\label{tab:offline_vqa_results}

\begin{tabular}{>{\raggedright\arraybackslash}m{2.2cm}cccccccc}
\toprule
\multirow[c]{3}{=}{\centering\arraybackslash\textbf{Method}} & \multicolumn{4}{c}{\textbf{MVCAP-Bench}} & \multicolumn{4}{c}{\textbf{MVCAP-Bench-FG}} \\
\cmidrule(lr){2-5} \cmidrule(lr){6-9}
& \textbf{L1} & \textbf{L2} & \textbf{L3} & \textbf{Avg.}
& \textbf{L1} & \textbf{L2} & \textbf{L3} & \textbf{Avg.} \\
\midrule
\rowcolor{black!8}\multicolumn{9}{l}{\textit{Offline}} \\
GPT-5-mini & 12.0 & 16.0 & \best{23.0} & 17.0 & 94.5 & 71.5 & 22.0 & 62.7 \\
GPT-5.4 & 17.0 & 18.0 & 22.0 & 19.0 & \best{100.0} & 72.0 & 27.0 & 66.3 \\
Claude-Sonnet-4-6 & 16.7 & 21.5 & 19.0 & 19.1 & \best{100.0} & 75.0 & 31.0 & 68.7 \\
Claude-Opus-4-6 & 16.5 & 21.0 & 21.5 & 19.7 & \best{100.0} & 76.5 & 32.5 & 69.7 \\
Gemini-3-Flash & \best{21.5} & \best{24.5} & 16.7 & \best{20.9} & \best{100.0} & \best{78.5} & 30.5 & 69.7 \\
Gemini-3.1-Pro & 18.1 & 19.0 & 20.5 & 19.2 & \best{100.0} & 77.0 & 33.0 & \best{70.0} \\
Doubao-Seed-2.0-Pro & 18.0 & 17.0 & 19.7 & 18.2 & \best{100.0} & 59.6 & 26.0 & 61.9 \\
GLM-4.6V & 19.0 & 13.0 & 21.0 & 17.7 & \best{100.0} & 70.5 & 21.5 & 64.0 \\
Kimi-K2.5 & 19.5 & 15.6 & 19.0 & 18.0 & \best{100.0} & 70.5 & 31.0 & 67.2 \\
Qwen3.5-Plus & 18.2 & 19.1 & 19.7 & 19.0 & \best{100.0} & 62.8 & \best{33.2} & 65.3 \\
\bottomrule
\end{tabular}
\end{table}

\begin{figure}[t]
    \centering
    \includegraphics[width=\linewidth]{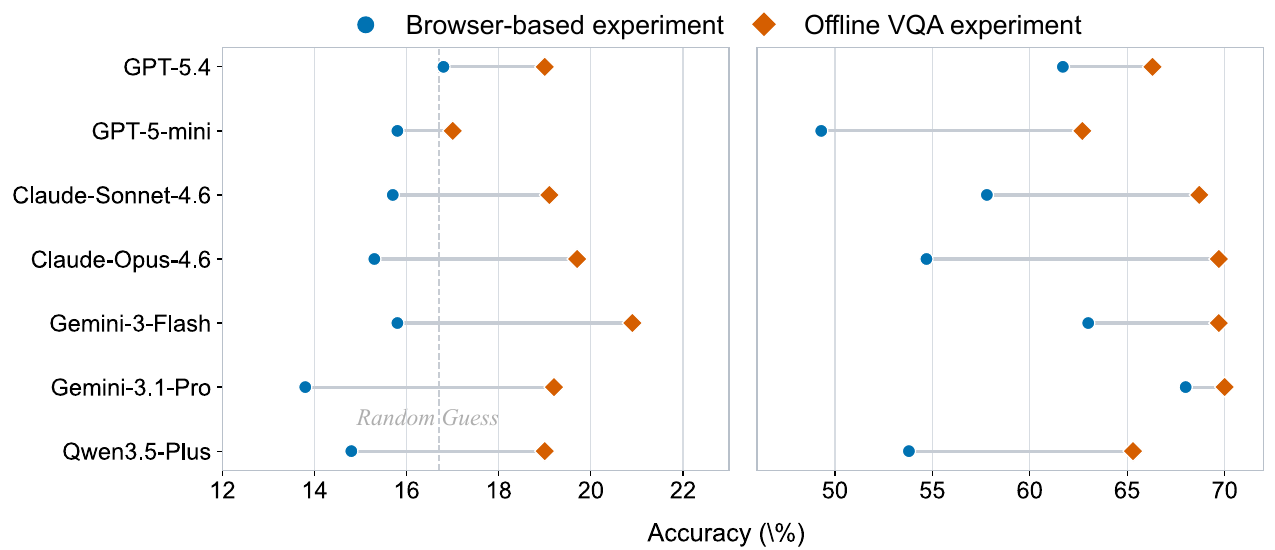}
    \caption{Comparison between browser-based and offline VQA experiments. Left: MVCAP-Bench. Right: MVCAP-Bench-FG. Offline gains are modest on the full benchmark but larger on the foreground-only control.} 
    \Description{A two-panel comparison of browser-based and offline VQA experiments for representative models on the same MVCAP instances. The left panel shows MVCAP-Bench and the right panel shows MVCAP-Bench-FG. Each row is one model, with a blue circle for the browser-based setting and an orange diamond for the offline VQA setting, connected by a line to indicate the change in average Pass@1. The figure shows smaller gains on the full benchmark and larger gains on the foreground-only setting.}
    \label{fig:offline_compare}
\end{figure}

The offline results are reported in Table~\ref{tab:offline_vqa_results}, and the comparison with browser-based performance is summarized in Fig.~\ref{fig:offline_compare}. Overall, the offline protocol helps clarify how much of the observed difficulty comes from browser interaction and how much persists as a failure to recover motion-defined semantics.

\noindent\textbf{Low Performance Persists in the Offline Protocol.}
Removing browser interaction yields only modest gains on the full MVCAP-Bench. Across the evaluated models, average accuracy increases from 13.8\%--16.8\% in the browser setting to 17.0\%--20.9\% offline, with Gemini-3-Flash reaching the best offline result of 20.9\%. Because performance remains close to the six-way baseline after interaction is removed, the observed difficulty cannot be attributed to browser control alone; under this protocol, recovering motion-defined semantics remains the main source of error.

\noindent\textbf{Browser interaction adds a secondary burden.}
The browser setting additionally requires webpage parsing, option grounding, context maintenance, and action execution. These demands are more visible on MVCAP-Bench-FG, where offline averages improve from 39.7\% to 64.0\% for GLM-4.6V, from 46.5\% to 67.2\% for Kimi-K2.5, and from 53.8\% to 65.3\% for Qwen3.5-Plus. On the full benchmark, offline gains are smaller and performance remains low in both settings. For the evaluated models, browser interaction is therefore a secondary source of error relative to recovering motion-defined semantics under full camouflage.

\section{Conclusion}

We presented \textbf{Motion Vision CAPTCHA (MVCAP)}, a hierarchical motion-based CAPTCHA framework that shifts challenge solving from static spatial recognition to motion-defined semantic perception. Built on coherent motion, structural motion, and biological motion, MVCAP exploits a fundamental asymmetry between human visual motion perception and current GUI agents. To evaluate this framework, we introduced \textbf{MVCAP-Bench} and its matched foreground-only control benchmark, \textbf{MVCAP-Bench-FG}. Extensive experiments in live browser, native computer use, and offline VQA settings reveal a large and consistent human--agent gap: humans solve the benchmark almost perfectly, whereas current agents remain near random guess on the full benchmark. These findings suggest that motion-defined perception remains a robust foundation for human--machine differentiation and a promising direction for future CAPTCHA design.

\begin{acks}
This work was supported in part by the National Natural Science Foundation of China under Grants 62522116, 62271312, 62132006, and 62225112, and in part by the Science and Technology Commission of Shanghai Municipality (STCSM) under Grant 22DZ2229005.
\end{acks}

\bibliographystyle{ACM-Reference-Format}
\bibliography{ref}

\end{document}